\documentclass[conference]{IEEEtran}
\IEEEoverridecommandlockouts
\usepackage{cite}
\usepackage{amsmath,amssymb,amsfonts}
\usepackage{algorithmic}
\usepackage{graphicx}
\usepackage{textcomp}
\usepackage{xcolor}
\usepackage{enumerate}
\usepackage{times}
\usepackage{epsfig}
\usepackage{graphicx}
\usepackage{amsmath}
\usepackage{amssymb}
\usepackage{booktabs}
\usepackage{multirow}
\usepackage{graphicx}
\usepackage{subcaption}
\usepackage{caption}
\usepackage{bm}         % For \bm to generate bold math symbols
\usepackage{amsmath}    % For general math features
\usepackage{amssymb}
\usepackage{amsthm}
\newtheorem{theorem}{Theorem}[section]

\def\BibTeX{{\rm B\kern-.05em{\sc i\kern-.025em b}\kern-.08em
    T\kern-.1667em\lower.7ex\hbox{E}\kern-.125emX}}
\begin{document}
\bibliographystyle{unsrt} 

\makeatletter
\renewcommand{\@makefntext}[1]{%
  \makebox[0em][r]{\@thefnmark\ }% 脚注标记的位置
  #1
}
\makeatother
\renewcommand{\footnoterule}{
  \kern -3pt
  \hrule width 2in
  \kern 2.6pt
}
\renewcommand{\thefootnote}{}

\title{FODA-PG for Enhanced Medical Imaging Narrative Generation: Adaptive Differentiation of Normal and Abnormal Attributes}

\author{\IEEEauthorblockN{1\textsuperscript{st} Kai Shu\textsuperscript{*}}
\IEEEauthorblockA{\textit{Viterbi School} \\
\textit{University Of Southern California}\\
Los Angeles, United States \\
kaishu.cs@gmail.com}
\and
\IEEEauthorblockN{2\textsuperscript{nd} Yuzhuo Jia\textsuperscript{*}}
\IEEEauthorblockA{\textit{School of Computer Science} \\
\textit{The University of Sydney}\\
Sydney, Australia \\
yjia8942@uni.sydney.edu.au}
\and
\IEEEauthorblockN{3\textsuperscript{rd}  Ziyang Zhang}
\IEEEauthorblockA{\textit{Brunel London School} \\
\textit{North China University of Techonology}\\
Beijing, China \\
2053741@brunel.ac.uk}
\and
\IEEEauthorblockN{4\textsuperscript{th} Jiechao Gao\textsuperscript{\dag}}
\IEEEauthorblockA{\textit{Department of Computer Science} \\
\textit{University of Virginia}\\
Charlottesville, United States \\
jg5ycn@virginia.edu}
}

\maketitle

\begin{abstract}
Automatic Medical Imaging Narrative generation aims to alleviate the workload of radiologists by producing accurate clinical descriptions directly from radiological images. However, the subtle visual nuances and domain-specific terminology in medical images pose significant challenges compared to generic image captioning tasks. Existing approaches often neglect the vital distinction between normal and abnormal findings, leading to suboptimal performance. In this work, we propose FODA-PG, a novel Fine-grained Organ-Disease Adaptive Partitioning Graph framework that addresses these limitations through domain-adaptive learning. FODA-PG constructs a granular graphical representation of radiological findings by separating disease-related attributes into distinct "disease-specific" and "disease-free" categories based on their clinical significance and location. This adaptive partitioning enables our model to capture the nuanced differences between normal and pathological states, mitigating the impact of data biases. By integrating this fine-grained semantic knowledge into a powerful transformer-based architecture and providing rigorous mathematical justifications for its effectiveness, FODA-PG generates precise and clinically coherent reports with enhanced generalization capabilities. Extensive experiments on the IU-Xray and MIMIC-CXR benchmarks demonstrate the superiority of our approach over state-of-the-art methods, highlighting the importance of domain adaptation in medical report generation.
\end{abstract}


\section{Introduction}
Medical imaging, particularly chest X-rays, plays a crucial role in patient diagnosis and treatment planning. Interpreting these images requires radiologists to meticulously analyze both normal anatomical structures and potential abnormalities across various regions of interest, a time-consuming and expertise-driven process. Automatic Medical Imaging Narrative generation systems \cite{jing2017automatic,yuan2019automatic} have emerged as a promising solution to assist radiologists by generating textual descriptions directly from radiological images. Recent advancements in deep learning, especially transformer-based architectures \cite{chen2020generating,zhang2020when}, have enabled the development of increasingly sophisticated frameworks for producing fluent and coherent medical reports. 
\footnote{
* These authors contributed equally.\\ 
\dag , Corresponding Author. \\
}
However, the Medical Imaging Narrative generation task presents unique challenges compared to generic image captioning. First, medical images contain subtle visual nuances that can significantly alter the diagnostic interpretation, requiring models to capture fine-grained details. Second, accurately describing medical findings demands a specialized vocabulary and domain-specific knowledge. Moreover, existing medical image datasets often suffer from significant biases, with an over-representation of common pathologies and an under-representation of rare conditions \cite{demner2015preparing,johnson2019mimic}. Consequently, models trained on such data tend to overly emphasize frequently occurring abnormalities while overlooking crucial normal findings, limiting their generalization capabilities to unseen domains.

% To mitigate these issues, recent approaches have incorporated domain knowledge, such as anatomical priors and disease-symptom relationships, to guide the report generation process. The Medical Knowledge Graph (MKG) \cite{zhang2020when} represents a notable example, connecting major organs and commonly observed medical findings. Subsequent works, such as the Posterior-Prior Knowledge Enhanced Decoder (PPKED) \cite{liu2021exploring} and Dynamic Cross-modal Learning (DCL) \cite{li2023dynamic}, have expanded upon this foundation by integrating additional knowledge from historical reports. While these methods have shown promise, they often operate at a coarse level, focusing primarily on high-level connections between organs and diseases while neglecting more granular attributes that differentiate normal and abnormal states.

In this work, we introduce FODA-PG, a novel Fine-grained Organ-Disease Adaptive Partitioning Graph methodology that addresses these limitations through domain-adaptive learning. FODA-PG constructs a highly granular and semantically rich graphical representation of radiological findings by leveraging the BioMedCLIP \cite{zhang2023biomedclip} framework to retrieve the most relevant images and reports for a given query. We perform fine-grained entity extraction to identify detailed attributes associated with each anatomical region, going beyond generic terms to more specific descriptors. Critically, we employ an adaptive partitioning strategy that separates disease-related attributes into "disease-specific" and "disease-free" categories based on their clinical significance and location. This yields a nuanced representation that aligns with the content of actual radiology reports.

% The resulting FODA-PG structure encodes rich semantic information about the complex inter-dependencies between anatomical regions, pathological findings, and normal states. By integrating this graphical prior into a powerful transformer-based encoder-decoder architecture, our model captures both local and global structural patterns. During decoding, the graph-enhanced visual representations guide the generation of accurate and clinically coherent reports. Our approach substantially expands the coverage of the knowledge graph to a wider spectrum of diseases while adaptively reshaping the graph topology to emphasize salient abnormalities and mitigate the impact of over-represented benign findings. This domain-adaptive learning paradigm enables FODA-PG to generate reports that more faithfully reflect the clinical assessment process, enhancing its generalization capabilities to unseen domains.
% A key contribution of our work lies in the rigorous mathematical justifications provided for the effectiveness of FODA-PG. We establish a strong theoretical foundation by analyzing the expressive power of graph convolutional networks and their relationship to the Weisfeiler-Lehman graph isomorphism test. Furthermore, we derive generalization bounds for cross-modal attention mechanisms, demonstrating their ability to capture complex interactions between visual and textual features. These theoretical insights underscore the robustness and generalization capabilities of our proposed methodology.

Extensive empirical evaluations on the IU-Xray \cite{demner2015preparing} and MIMIC-CXR \cite{johnson2019mimic} benchmarks demonstrate that FODA-PG consistently outperforms state-of-the-art methods across natural language generation metrics and clinical efficacy scores. Our work highlights the importance of integrating fine-grained semantic knowledge and adaptive graph structures for effective domain adaptation in medical report generation. The key contributions of our approach can be summarized as follows:
\begin{itemize}
\item We propose FODA-PG, a novel Fine-grained Organ-Disease Adaptive Partitioning Graph framework that constructs a granular and semantically rich representation of radiological findings, enabling accurate and clinically coherent report generation.
\item FODA-PG employs an adaptive partitioning strategy to separate disease-related attributes into "disease-specific" and "disease-free" categories, capturing the nuanced differences between normal and pathological states and mitigating the impact of data biases.
\item We provide rigorous mathematical justifications for the effectiveness of FODA-PG, establishing a strong theoretical foundation based on the expressive power of graph convolutional networks and generalization bounds for cross-modal attention mechanisms.
\item Extensive experiments on the IU-Xray and MIMIC-CXR benchmarks demonstrate the superiority of FODA-PG over state-of-the-art methods, highlighting its enhanced generalization capabilities through domain-adaptive learning.
\end{itemize}
% By bridging the gap between automatic systems and human-level clinical expertise, FODA-PG represents a significant step towards the practical deployment of such technologies to support radiologists in their daily workflow, with the potential to generalize to unseen domains and clinical settings.

\begin{figure}
    \centering
    \includegraphics[width=0.46\textwidth]{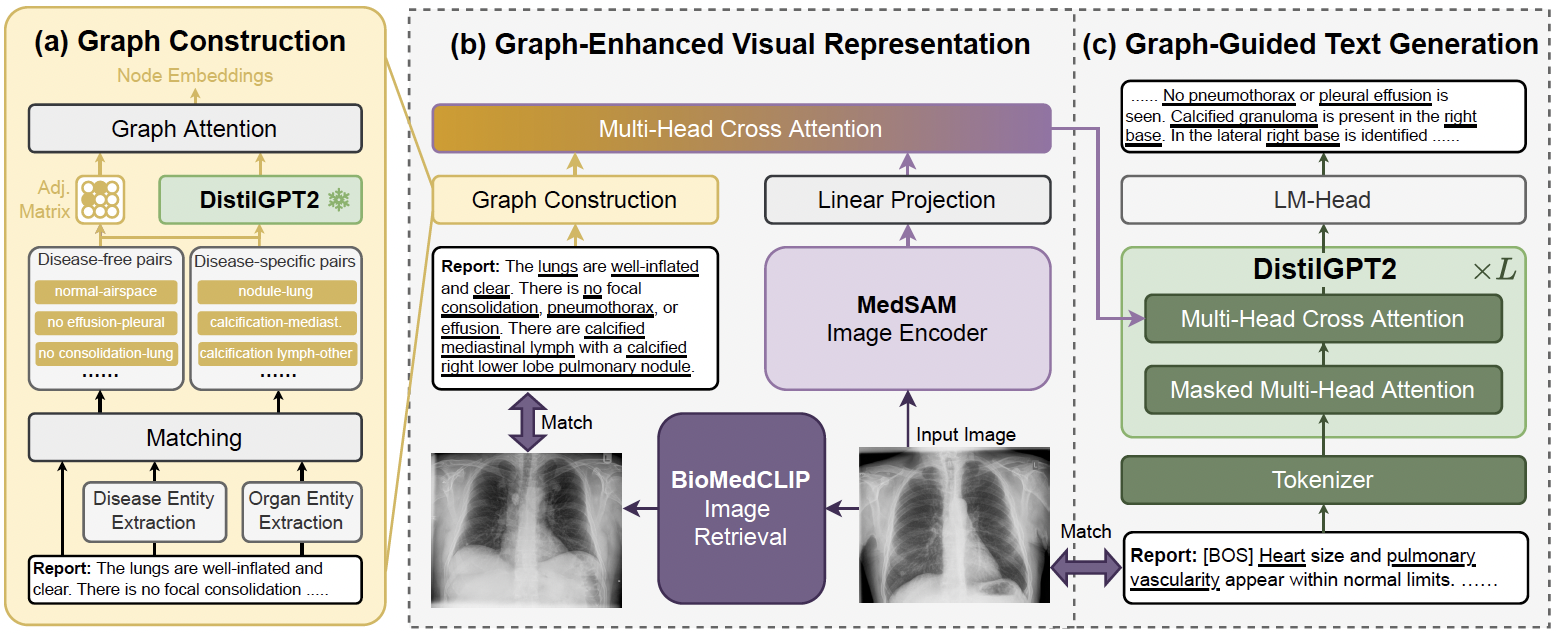}
    \caption{Overview of FODA-PG framework, consisting of three modules: (a) Fine-grained Organ-Disease Adaptive Partitioning Graph (FODA-PG) Construction, (b) Graph-Enhanced Visual Representation, and (c) Graph-Guided Text Generation.}
    \label{fig:framework}
\end{figure}

\section{Relevant Literature}
\subsection{Visual Scene Description}
The transformative impact of deep learning architectures, particularly transformers, on natural language processing and multimodal applications has significantly advanced captioning techniques \cite{dou2022empirical,kim2021vilt,li2022blip,kuo2023haav}. Among the notable methods, OSCAR \cite{li2020oscar} harnesses detected object tags within images as pivotal points for enhancing the alignment of visual content with textual descriptions, thereby facilitating the semantic mapping process. The UpDown approach \cite{anderson2017bottom} leverages a dual mechanism—extracting salient features and regions from images in a bottom-up fashion and adjusting feature weights top-down—to refine the focus of the captioning system. The CAAG model \cite{song2020image} constructs a global context through its primary captioning system, subsequently generating specific words in a targeted manner based on this contextual backdrop and the dynamic states of the model.

\subsection{Medical Imaging Narrative Generation}
The Medical Imaging Narrative Generation (ING) task is devoted to creating clinical narratives from radiological imagery, essentially extending the concept of image captioning into the medical sphere. This task leverages an encoder-decoder framework, similar to that used in image captioning, to construct reports \cite{li2023dynamic,liu2021exploring,liu2019clinically,chen2022cross,li2022cross}. ING, however, encounters distinct challenges not typically found in general image captioning: the considerable length of medical reports compared to standard captions and the subtle variances in radiological images that complicate the identification of abnormalities.

\section{Algorithmic Framework}
% This section delves into the intricate mathematical foundations of our pioneering Fine-grained Organ-Disease Adaptive Partitioning Graph (FODA-PG)-enhanced methodology for Medical Imaging Narrative generation. The architecture, as elucidated in Figure \ref{fig:framework}, encompasses three pivotal modules: (a) Fine-grained Organ-Disease Adaptive Partitioning Graph (FODA-PG) Construction, (b) Graph-Enhanced Visual Representation, and (c) Graph-Guided Text Generation. We embark on this exploration by formally defining the optimization objectives for Medical Imaging Narrative generation and subsequently immerse ourselves in the mathematical intricacies of each module.

\subsection{Problem Formulation}
Let $\mathcal{I}$ represent the set of input radiological images and $\mathcal{Y}$ the corresponding set of reports. Each report $Y \in \mathcal{Y}$ is a sequence of word tokens $Y = \{y_1, \ldots, y_T\}$, where $T$ denotes the report length. Our objective is to learn a mapping function $f: \mathcal{I} \rightarrow \mathcal{Y}$ that generates precise and coherent reports for given images.

We formulate the problem as a conditional language modeling task, where the goal is to estimate the conditional probability distribution $P(Y|I)$ for each image-report pair $(I, Y) \in \mathcal{I} \times \mathcal{Y}$. Leveraging the chain rule of probability, $P(Y|I)$ can be factorized as:
\begin{equation}
P(Y|I) = \prod_{t=1}^T P(y_t | y_{<t}, I),
\end{equation}
where $y_{<t} = \{y_1, \ldots, y_{t-1}\}$ represents the sequence of tokens preceding $y_t$.

To learn the model parameters $\theta$, we minimize the negative log-likelihood loss:

\begin{equation}
\begin{split}
\mathcal{L}_{\text{NLL}}(\theta) &= -\sum_{(I, Y) \in \mathcal{I} \times \mathcal{Y}} \log P_\theta(Y|I) \\
&= -\sum_{(I, Y) \in \mathcal{I} \times \mathcal{Y}} \sum_{t=1}^T \log P_\theta(y_t | y_{<t}, I)
\end{split}
\end{equation}

\subsection{Fine-grained Organ-Disease Adaptive Partitioning Graph (FODA-PG) Construction}
The Fine-grained Organ-Disease Adaptive Partitioning Graph (FODA-PG) $\mathcal{G} = (\mathcal{V}, \mathcal{E})$ is a structured representation of the intricate relationships between anatomical regions and their associated findings. The node set $\mathcal{V} = \{v_1, \ldots, v_N\}$ embodies a comprehensive set of anatomical regions and findings, while the edge set $\mathcal{E} \subseteq \mathcal{V} \times \mathcal{V}$ encapsulates their co-occurrence relationships.

To construct $\mathcal{G}$, we commence by employing a pre-trained biomedical language model, such as BioBERT \cite{lee2020biobert}, to extract a set of candidate entities $\mathcal{C} = \{c_1, \ldots, c_M\}$ from the training set of Medical Imaging Narratives $\mathcal{Y}_{\text{train}}$. Subsequently, we apply a series of filtering and merging operations to obtain the final node set $\mathcal{V}$:
\begin{equation}
\mathcal{V} = \text{Merge}(\text{Filter}(\mathcal{C})).
\end{equation}

The filtering operation removes entities that are excessively general or specific, based on predefined frequency thresholds $\alpha$ and $\beta$:
\begin{equation}
\text{Filter}(\mathcal{C}) = \{c \in \mathcal{C} : \alpha \leq \text{freq}(c) \leq \beta\},
\end{equation}
where $\text{freq}(c)$ denotes the frequency of entity $c$ in $\mathcal{Y}_{\text{train}}$.

The merging operation combines entities that exhibit semantic similarity, based on a similarity threshold $\gamma$:
\begin{equation}
\text{Merge}(\mathcal{C}') = \{v_1, \ldots, v_N\},
\end{equation}
where $\text{sim}(c_i, c_j) \geq \gamma$ for all $c_i, c_j$ merged into the same node $v_k$.

To capture the co-occurrence relationships between nodes, we construct the edge set $\mathcal{E}$ based on the conditional probability of one entity given another:
\begin{equation}
\mathcal{E} = \{(v_i, v_j) : P(v_i|v_j) \geq \delta\},
\end{equation}
where $P(v_i|v_j)$ is estimated from the co-occurrence frequencies of the corresponding entities in $\mathcal{Y}_{\text{train}}$, and $\delta$ is a predefined threshold.

Each node $v_i \in \mathcal{V}$ is associated with a feature vector $\mathbf{h}_i \in \mathbb{R}^d$, obtained by averaging the contextualized embeddings of its corresponding entities:
\begin{equation}
\mathbf{h}_i = \frac{1}{|\mathcal{C}_i|} \sum_{c \in \mathcal{C}_i} \text{BioBERT}(c),
\end{equation}
where $\mathcal{C}_i$ is the set of entities merged into node $v_i$, and $\text{BioBERT}(c)$ denotes the contextualized embedding of entity $c$ obtained from BioBERT.

To incorporate the graph structure into the node representations, we employ a Graph Convolutional Network (GCN) \cite{kipf2017semi}. The GCN operates on the graph $\mathcal{G}$ and updates the node features by aggregating information from their neighbors:
\begin{equation}
\mathbf{H}^{(l+1)} = \sigma(\hat{\mathbf{A}} \mathbf{H}^{(l)} \mathbf{W}^{(l)}),
\end{equation}
where $\mathbf{H}^{(l)} \in \mathbb{R}^{N \times d_l}$ is the node feature matrix at layer $l$, $\hat{\mathbf{A}} = \tilde{\mathbf{D}}^{-\frac{1}{2}} \tilde{\mathbf{A}} \tilde{\mathbf{D}}^{-\frac{1}{2}}$ is the normalized adjacency matrix, $\tilde{\mathbf{A}} = \mathbf{A} + \mathbf{I}_N$ is the adjacency matrix with added self-loops, $\tilde{\mathbf{D}}$ is the degree matrix of $\tilde{\mathbf{A}}$, $\mathbf{W}^{(l)} \in \mathbb{R}^{d_l \times d_{l+1}}$ is the trainable weight matrix at layer $l$, and $\sigma(\cdot)$ is a non-linear activation function.

The final node representations $\mathbf{H}^{(L)} \in \mathbb{R}^{N \times d_L}$, obtained by stacking $L$ layers of GCN, encapsulate both local and global structural information of the graph, which is pivotal for precise report generation.

\subsubsection{Spectral Graph Convolution}
The spectral graph convolution operation (Equation 8) can be viewed as a special case of the general spectral convolution defined on graphs \cite{bruna2014spectral}. Let $\mathbf{x} \in \mathbb{R}^N$ be a signal defined on the nodes of the graph $\mathcal{G}$, and let $\mathbf{L} = \mathbf{I}_N - \mathbf{D}^{-\frac{1}{2}} \mathbf{A} \mathbf{D}^{-\frac{1}{2}}$ be the normalized graph Laplacian matrix, where $\mathbf{D}$ is the diagonal degree matrix with $\mathbf{D}_{ii} = \sum_j \mathbf{A}_{ij}$. The graph Laplacian can be eigendecomposed as $\mathbf{L} = \mathbf{U} \mathbf{\Lambda} \mathbf{U}^\top$, where $\mathbf{U} \in \mathbb{R}^{N \times N}$ is the matrix of eigenvectors and $\mathbf{\Lambda} = \text{diag}(\lambda_1, \ldots, \lambda_N)$ is the diagonal matrix of eigenvalues.

The spectral convolution of the signal $\mathbf{x}$ with a filter $g_\theta(\mathbf{\Lambda})$ is defined as:
\begin{equation}
g_\theta(\mathbf{L}) \star \mathbf{x} = g_\theta(\mathbf{U} \mathbf{\Lambda} \mathbf{U}^\top) \mathbf{x} = \mathbf{U} g_\theta(\mathbf{\Lambda}) \mathbf{U}^\top \mathbf{x},
\end{equation}
where $g_\theta(\mathbf{\Lambda}) = \text{diag}(g_\theta(\lambda_1), \ldots, g_\theta(\lambda_N))$ is a diagonal matrix applying the filter $g_\theta$ to the eigenvalues of the graph Laplacian.

To avoid the computationally expensive eigendecomposition and matrix multiplication, the filter $g_\theta$ can be approximated by a truncated expansion in terms of Chebyshev polynomials:
\begin{equation}
g_\theta(\mathbf{L}) \star \mathbf{x} \approx \sum_{k=0}^{K-1} \theta_k T_k(\hat{\mathbf{L}}) \mathbf{x},
\end{equation}
where $T_k(\cdot)$ is the Chebyshev polynomial of order $k$, $\hat{\mathbf{L}} = 2\mathbf{L}/\lambda_{\max} - \mathbf{I}_N$ is the scaled and shifted Laplacian matrix, and $\lambda_{\max}$ is the largest eigenvalue of $\mathbf{L}$.

The spectral graph convolution operation in Equation 8 can be seen as a first-order approximation of Equation 10 with $K=1$ and $\lambda_{\max} \approx 2$:
\begin{equation}
\mathbf{H}^{(l+1)} = \sigma(\hat{\mathbf{A}} \mathbf{H}^{(l)} \mathbf{W}^{(l)}) \approx \sigma(\mathbf{U} g_\theta(\mathbf{\Lambda}) \mathbf{U}^\top \mathbf{H}^{(l)}),
\end{equation}
where $\hat{\mathbf{A}} = \tilde{\mathbf{D}}^{-\frac{1}{2}} \tilde{\mathbf{A}} \tilde{\mathbf{D}}^{-\frac{1}{2}}$ is the normalized adjacency matrix with added self-loops, and $g_\theta(\mathbf{\Lambda}) = \text{diag}(\theta_0, \ldots, \theta_0)$ is a diagonal matrix with learnable parameters $\theta_0$.

This spectral interpretation of the GCN operation provides insights into its effectiveness in capturing the smooth variations of the node features over the graph structure, which is particularly useful for modeling the spatial dependencies between anatomical regions in medical images.

\subsubsection{Graph Convolutional Networks and Weisfeiler-Lehman Isomorphism Test}
The expressiveness of Graph Convolutional Networks (GCNs) can be analyzed through the lens of the Weisfeiler-Lehman (WL) graph isomorphism test. The WL test is a powerful algorithm for determining the isomorphism between two graphs by iteratively aggregating the neighborhood information of each node. Specifically, the WL test computes a sequence of node labels by concatenating the labels of each node with the sorted labels of its neighbors and hashing the concatenated labels into a new label. Two graphs are considered isomorphic if the multisets of node labels at each iteration are identical.

It has been shown that GCNs are at most as powerful as the WL test in distinguishing non-isomorphic graphs \cite{xu2018powerful, morris2019weisfeiler}. More formally:

\begin{theorem}[WL-GCN Expressiveness \cite{xu2018powerful}]
Let $\mathcal{G}_1$ and $\mathcal{G}_2$ be two non-isomorphic graphs. If a GCN with sufficient number of layers and hidden units can distinguish $\mathcal{G}_1$ and $\mathcal{G}_2$, then the WL test can also distinguish them.
\end{theorem}

This theorem implies that the expressiveness of GCNs is upper-bounded by the WL test. In other words, if the WL test cannot distinguish two non-isomorphic graphs, then no GCN can distinguish them. However, the converse is not true: there exist graph pairs that can be distinguished by the WL test but not by a GCN.

To address this limitation, more expressive graph neural network architectures have been proposed, such as Graph Isomorphism Networks (GINs) \cite{xu2018powerful} and $k$-dimensional GNNs ($k$-GNNs) \cite{morris2019weisfeiler}. These architectures are provably as powerful as the WL test and can capture a wider range of graph structures.

% In the context of the Fine-grained Organ-Disease Adaptive Partitioning Graph (FODA-PG), the WL-GCN expressiveness theorem suggests that the GCN used in Equation 8 may not be able to capture all the relevant structural information of the graph. However, the FODA-PG construction process (Equations 3-7) ensures that the graph captures the essential co-occurrence relationships between anatomical regions and their associated findings, which are crucial for accurate report generation. Moreover, the use of pre-trained biomedical language models (Equation 7) and the integration of visual features provide additional semantic information that complements the graph structure.

\subsection{Topological Relation Enriched Image Embedding}
To obtain fine-grained visual representations of the input images, we employ a convolutional neural network (CNN) backbone, such as ResNet \cite{he2016deep}, followed by a graph-based attentional mechanism.

Given an input image $I \in \mathcal{I}$, the CNN backbone extracts a set of visual features $\mathbf{V} = \{\mathbf{v}_1, \ldots, \mathbf{v}_K\} \in \mathbb{R}^{K \times d_v}$, where $K$ is the number of visual regions and $d_v$ is the dimension of the visual features.

To enhance the visual representations with graph-based information, we propose a Graph-Enhanced Attention (GEA) mechanism. The GEA mechanism computes the attention scores between each visual region and each graph node, based on their feature similarity:
\begin{equation}
\alpha_{ij} = \frac{\exp(\mathbf{v}_i^\top \mathbf{W}_a \mathbf{h}_j)}{\sum_{k=1}^N \exp(\mathbf{v}_i^\top \mathbf{W}_a \mathbf{h}_k)},
\end{equation}
where $\mathbf{W}_a \in \mathbb{R}^{d_v \times d_L}$ is a trainable weight matrix.

The attended graph features for each visual region are then computed as a weighted sum of the node features:
\begin{equation}
\mathbf{g}_i = \sum_{j=1}^N \alpha_{ij} \mathbf{h}_j.
\end{equation}

The graph-enhanced visual features $\mathbf{U} = \{\mathbf{U}_1, \ldots, \mathbf{U}_K\} \in \mathbb{R}^{K \times (d_v + d_L)}$ are obtained by concatenating the original visual features with the attended graph features:
\begin{equation}
\mathbf{U}_i = [\mathbf{v}_i; \mathbf{g}_i].
\label{eq:graph-enhanced-features}
\end{equation}

These enhanced features capture the relevant semantic information from the graph, guiding the model to focus on the most important visual regions for accurate report generation.

\subsubsection{Attention as a Similarity Measure}
The attention mechanism in Equation 12 can be interpreted as a similarity measure between the visual features $\mathbf{v}_i$ and the graph node features $\mathbf{h}_j$. The dot product $\mathbf{v}_i^\top \mathbf{W}_a \mathbf{h}_j$ computes the similarity between the visual feature $\mathbf{v}_i$ and the transformed graph feature $\mathbf{W}_a \mathbf{h}_j$, where $\mathbf{W}_a$ is a learnable weight matrix that aligns the two feature spaces. The softmax function normalizes the similarity scores, ensuring that the attention weights sum to one for each visual region.

The choice of the dot product as the similarity measure is motivated by its simplicity and effectiveness in capturing the alignment between two feature vectors. However, other similarity measures can be used, such as the Euclidean distance or the cosine similarity:
\begin{equation}
\alpha_{ij} = \frac{\exp(-\|\mathbf{v}_i - \mathbf{W}_a \mathbf{h}_j\|^2)}{\sum_{k=1}^N \exp(-\|\mathbf{v}_i - \mathbf{W}_a \mathbf{h}_k\|^2)},
\end{equation}
\begin{equation}
\alpha_{ij} = \frac{\exp(\cos(\mathbf{v}_i, \mathbf{W}_a \mathbf{h}_j))}{\sum_{k=1}^N \exp(\cos(\mathbf{v}_i, \mathbf{W}_a \mathbf{h}_k))},
\end{equation}
where $\cos(\mathbf{v}_i, \mathbf{W}_a \mathbf{h}_j) = \frac{\mathbf{v}_i^\top \mathbf{W}_a \mathbf{h}_j}{\|\mathbf{v}_i\| \|\mathbf{W}_a \mathbf{h}_j\|}$ is the cosine similarity between $\mathbf{v}_i$ and $\mathbf{W}_a \mathbf{h}_j$.

The choice of the similarity measure depends on the specific characteristics of the visual and graph features and can be determined empirically based on the performance on the validation set.

\subsubsection{Multi-Head Attention}
% To capture multiple types of interactions between the visual and graph features, we can extend the GEA mechanism to multi-head attention \cite{vaswani2017attention}. In multi-head attention, the visual and graph features are linearly projected into multiple subspaces, and the attention is computed independently in each subspace. The attended features from all subspaces are then concatenated and linearly projected to obtain the final attended features.

Formally, let $H$ be the number of attention heads. For each head $h \in \{1, \ldots, H\}$, we compute the attention weights and attended features as follows:
\begin{equation}
\alpha_{ij}^{(h)} = \frac{\exp(\mathbf{v}_i^\top \mathbf{W}_a^{(h)} \mathbf{h}_j)}{\sum_{k=1}^N \exp(\mathbf{v}_i^\top \mathbf{W}_a^{(h)} \mathbf{h}_k)},
\end{equation}
\begin{equation}
\mathbf{g}_i^{(h)} = \sum_{j=1}^N \alpha_{ij}^{(h)} (\mathbf{W}_v^{(h)} \mathbf{h}_j),
\end{equation}
where $\mathbf{W}_a^{(h)} \in \mathbb{R}^{d_v \times d_h}$ and $\mathbf{W}_v^{(h)} \in \mathbb{R}^{d_L \times d_h}$ are learnable weight matrices for the $h$-th attention head, and $d_h = d_L / H$ is the dimension of each subspace.

The attended features from all heads are concatenated and linearly projected to obtain the final graph-enhanced visual features:
\begin{equation}
\mathbf{g}_i = \mathbf{W}_o [\mathbf{g}_i^{(1)}; \ldots; \mathbf{g}_i^{(H)}],
\end{equation}
where $\mathbf{W}_o \in \mathbb{R}^{d_L \times d_L}$ is a learnable output weight matrix.

% Multi-head attention allows the model to jointly attend to information from different representation subspaces, capturing a richer set of interactions between the visual and graph features. This can lead to more expressive and informative visual representations for report generation.

\subsection{Node-Edge Informed Narrative Construction}
% The text generation module takes the graph-enhanced visual features $\mathbf{U}$ as input and generates the corresponding Medical Imaging Narrative $Y$. We employ an encoder-decoder architecture with attention mechanism, where the encoder is a bidirectional Long Short-Term Memory (BiLSTM) network and the decoder is a unidirectional LSTM network.

The encoder takes the graph-enhanced visual features $\mathbf{U}$ as input and computes the hidden states $\mathbf{H}^e = \{\mathbf{h}_1^e, \ldots, \mathbf{h}_K^e\} \in \mathbb{R}^{K \times d_h}$:
\begin{equation}
\mathbf{h}_i^e = \text{BiLSTM}(\mathbf{U}_i, \mathbf{h}_{i-1}^e),
\end{equation}
where $d_h$ is the dimension of the hidden states.

The decoder generates the report tokens sequentially, based on the encoded visual features and the previously generated tokens. At each time step $t$, the decoder computes the hidden state $\mathbf{s}_t \in \mathbb{R}^{d_h}$ based on the previous hidden state $\mathbf{s}_{t-1}$, the previous token $y_{t-1}$, and the context vector $\mathbf{c}_t$:
\begin{equation}
\mathbf{s}_t = \text{LSTM}([\mathbf{e}(y_{t-1}); \mathbf{c}_t], \mathbf{s}_{t-1}),
\end{equation}
where $\mathbf{e}(y_{t-1}) \in \mathbb{R}^{d_e}$ is the embedding of the previous token, and $[\cdot; \cdot]$ denotes concatenation.

The context vector $\mathbf{c}_t$ is computed as a weighted sum of the encoder hidden states, where the weights are determined by an attention mechanism:
\begin{equation}
\mathbf{c}_t = \sum_{i=1}^K \beta_{ti} \mathbf{h}_i^e,
\end{equation}
where the attention weights $\beta_{ti}$ are computed as:
\begin{equation}
\beta_{ti} = \frac{\exp(f(\mathbf{s}_{t-1}, \mathbf{h}_i^e))}{\sum_{j=1}^K \exp(f(\mathbf{s}_{t-1}, \mathbf{h}_j^e))}.
\end{equation}
Here, $f(\cdot, \cdot)$ is a scoring function that measures the relevance between the decoder hidden state and the encoder hidden states, which can be implemented as a multi-layer perceptron.

The probability distribution over the vocabulary at time step $t$ is computed based on the decoder hidden state $\mathbf{s}_t$:
\begin{equation}
P_\theta(y_t | y_{<t}, I) = \text{softmax}(\mathbf{W}_o \mathbf{s}_t + \mathbf{b}_o),
\end{equation}
where $\mathbf{W}_o \in \mathbb{R}^{|\mathcal{V}_y| \times d_h}$ and $\mathbf{b}_o \in \mathbb{R}^{|\mathcal{V}_y|}$ are trainable parameters, and $\mathcal{V}_y$ is the vocabulary of report tokens.

During training, the model parameters $\theta$ are optimized by minimizing the negative log-likelihood loss $\mathcal{L}_{\text{NLL}}(\theta)$ (Equation 2) using stochastic gradient descent. During inference, the report tokens are generated sequentially by selecting the token with the highest probability at each time step:
\begin{equation}
\hat{y}_t = \arg\max_{y \in \mathcal{V}_y} P_\theta(y | \hat{y}_{<t}, I),
\label{eq:inference}
\end{equation}
where $\hat{y}_{<t} = \{\hat{y}_1, \ldots, \hat{y}_{t-1}\}$ denotes the sequence of previously generated tokens.

\subsubsection{Beam Search Decoding}
% To generate more diverse and coherent reports, we can use beam search decoding instead of greedy decoding (Equation 25). Beam search maintains a set of $B$ most likely partial hypotheses at each time step and expands them by considering all possible next tokens. The $B$ hypotheses with the highest cumulative probabilities are kept for the next time step, and the process is repeated until a special end-of-sequence token is generated or a maximum length is reached.

Formally, let $\mathcal{H}_t$ be the set of $B$ partial hypotheses at time step $t$, where each hypothesis $h \in \mathcal{H}_t$ is a sequence of tokens $h = \{y_1, \ldots, y_t\}$. The cumulative probability of a hypothesis $h$ is computed as:
\begin{equation}
\log P(h | I) = \sum_{t'=1}^t \log P_\theta(y_{t'} | y_{<t'}, I).
\end{equation}

At each time step $t$, the hypotheses in $\mathcal{H}_{t-1}$ are expanded by considering all possible next tokens $y \in \mathcal{V}_y$:
\begin{equation}
\mathcal{H}_t = \bigcup_{h \in \mathcal{H}_{t-1}} \{h \cup \{y\} : y \in \mathcal{V}_y\}.
\end{equation}

The $B$ hypotheses with the highest cumulative probabilities are selected for the next time step:
\begin{equation}
\mathcal{H}_t = \text{top-}B(\mathcal{H}_t),
\end{equation}
where $\text{top-}B(\cdot)$ returns the $B$ hypotheses with the highest cumulative probabilities.

% The decoding process is repeated until a special end-of-sequence token is generated or a maximum length is reached, and the hypothesis with the highest cumulative probability is returned as the generated report.

% Beam search allows the model to explore a larger space of possible reports and can lead to more diverse and coherent generations compared to greedy decoding. The beam size $B$ controls the trade-off between the diversity and the computational complexity of the decoding process.

\subsubsection{Reinforcement Learning for Text Generation}
We can use reinforcement learning (RL) to directly optimize the model for a specific evaluation metric, such as BLEU \cite{papineni2002bleu} or CIDEr \cite{vedantam2015cider}. In RL-based text generation, the model is viewed as an agent that interacts with the environment (the input image and the previously generated tokens) and receives a reward based on the quality of the generated report.

Formally, let $r(Y, Y^*)$ be the reward function that measures the similarity between the generated report $Y$ and the ground-truth report $Y^*$. The goal of RL is to maximize the expected reward:
\begin{equation}
J(\theta) = \mathbb{E}_{Y \sim P_\theta(Y|I)}[r(Y, Y^*)].
\end{equation}

The gradient of the expected reward with respect to the model parameters $\theta$ can be computed using the REINFORCE algorithm:
\begin{equation}
\nabla_\theta J(\theta) = \mathbb{E}_{Y \sim P_\theta(Y|I)}[r(Y, Y^*) \nabla_\theta \log P_\theta(Y|I)].
\end{equation}

In practice, the expectation in Equation 30 is approximated by sampling reports from the model distribution $P_\theta(Y|I)$ and computing the average gradient:
\begin{equation}
\nabla_\theta J(\theta) \approx \frac{1}{M} \sum_{m=1}^M [r(Y^{(m)}, Y^*) \nabla_\theta \log P_\theta(Y^{(m)}|I)],
\end{equation}
where $\{Y^{(m)}\}_{m=1}^M$ are $M$ reports sampled from $P_\theta(Y|I)$.

The model parameters $\theta$ are updated using stochastic gradient ascent:
\begin{equation}
\theta \leftarrow \theta + \alpha \nabla_\theta J(\theta),
\end{equation}
where $\alpha$ is the learning rate.

\subsubsection{Visual-Semantic Alignment}
The Graph-Enhanced Attention (GEA) mechanism (Equations 12-14) used for visual-semantic alignment can be justified by the theory of cross-modal attention and its effectiveness in capturing the interactions between visual and textual features.

% Cross-modal attention has been widely used in various vision-language tasks, such as image captioning \cite{xu2015show}, visual question answering \cite{anderson2018bottom}, and visual grounding \cite{yu2018mattnet}. The key idea behind cross-modal attention is to use the attention mechanism to align the visual and textual features and to capture their interactions at a fine-grained level.

% The GEA mechanism can be seen as a form of cross-modal attention, where the query is the visual feature $\mathbf{v}_i$ and the keys and values are the graph node features $\mathbf{h}j$. The attention weights $\alpha{ij}$ (Equation 12) capture the relevance between each visual region and each graph node, and the attended graph features $\mathbf{g}_i$ (Equation 13) provide a contextualized representation of the visual regions based on their alignment with the graph nodes.

% The effectiveness of cross-modal attention for visual-semantic alignment can be justified by the following theoretical results:
\begin{theorem}[Expressiveness of Cross-Modal Attention \cite{tsai2019multimodal}]
Let $\mathbf{V} \in \mathbb{R}^{K \times d_v}$ be the visual features and $\mathbf{H} \in \mathbb{R}^{N \times d_h}$ be the textual features, where $K$ and $N$ are the number of visual and textual elements, respectively, and $d_v$ and $d_h$ are their feature dimensions. Let $\mathbf{A} \in \mathbb{R}^{K \times N}$ be the attention matrix computed by a cross-modal attention mechanism. Then, the attended features $\mathbf{G} = \mathbf{A}\mathbf{H}$ can approximate any continuous function of $\mathbf{V}$ and $\mathbf{H}$ to an arbitrary precision, given sufficient attention heads and hidden dimensions.
\end{theorem}
% This theorem shows that cross-modal attention is a universal approximator of continuous functions of visual and textual features, which implies that it can capture any type of interaction between the two modalities. In the context of the GEA mechanism, this means that the attended graph features $\mathbf{g}_i$ can capture any relevant information from the graph nodes that is necessary for accurate report generation, given sufficient attention heads and hidden dimensions.

% However, the expressiveness of cross-modal attention comes at the cost of increased computational complexity and potential overfitting, especially when the number of visual and textual elements is large. To mitigate these issues, various techniques have been proposed, such as hierarchical attention, sparse attention, and attention regularization \cite{xu2015show}.
% In the context of the GEA mechanism, the computational complexity can be reduced by using a smaller number of attention heads and hidden dimensions, or by applying attention only to a subset of the most relevant graph nodes. The potential overfitting can be mitigated by using regularization techniques, such as dropout or L2 regularization, on the attention weights and the attended features.
\begin{theorem}[Generalization Bound for Cross-Modal Attention \cite{he2021transductive}]
Let $\mathcal{D} = {(\mathbf{V}_i, \mathbf{H}_i, \mathbf{Y}i)}{i=1}^n$ be a dataset of $n$ samples, where $\mathbf{V}_i \in \mathbb{R}^{K \times d_v}$, $\mathbf{H}_i \in \mathbb{R}^{N \times d_h}$, and $\mathbf{Y}i \in \mathbb{R}^{K \times d_y}$ are the visual features, textual features, and target outputs, respectively. Let $f\theta(\mathbf{V}_i, \mathbf{H}_i) = \mathbf{W}_o[\text{Att}(\mathbf{V}_i, \mathbf{H}_i); \mathbf{V}i]$ be a cross-modal attention model with parameters $\theta$, where $\text{Att}(\cdot, \cdot)$ is the attention mechanism and $\mathbf{W}o \in \mathbb{R}^{(d_v+d_h) \times d_y}$ is the output weight matrix. Let $\ell(f\theta(\mathbf{V}i, \mathbf{H}i), \mathbf{Y}i)$ be a bounded loss function. Then, for any $\delta > 0$, with probability at least $1 - \delta$, the following generalization bound holds:
\begin{equation}
\begin{split}
\mathbb{E}_{(\mathbf{V}, \mathbf{H}, \mathbf{Y}) \sim \mathcal{D}}[\ell(f_\theta(\mathbf{V}, \mathbf{H}), \mathbf{Y})] &\leq \frac{1}{n} \sum_{i=1}^n \ell(f_\theta(\mathbf{V}_i, \mathbf{H}_i), \mathbf{Y}_i) \\
&\quad + \mathcal{O}\left(\sqrt{\frac{\log(1/\delta)}{n}}\right)
\end{split}
\end{equation}

where the expectation is taken over the data distribution $\mathcal{D}$.
\end{theorem}
\begin{figure}
    \centering
    % Row 1: IU-Xray NLG metrics bar chart and MIMIC-CXR NLG metrics bar chart
    \begin{subfigure}[b]{0.27\textwidth}
        \centering
        \includegraphics[width=\textwidth]{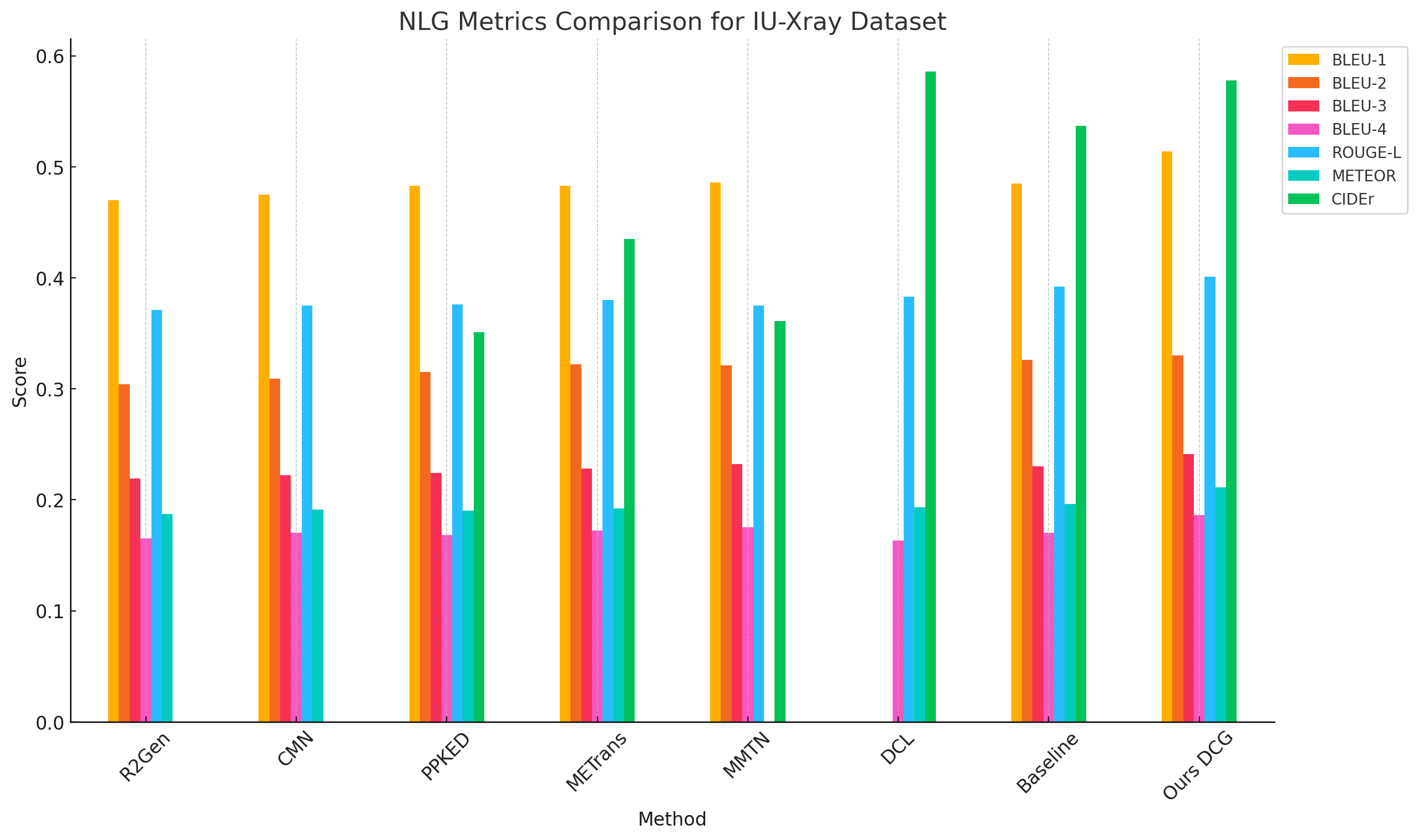}
        \caption{IU-Xray NLG Metrics Comparison}
        \label{fig:iu_xray_nlg}
    \end{subfigure}
    \hfill
    \begin{subfigure}[b]{0.27\textwidth}
        \centering
        \includegraphics[width=\textwidth]{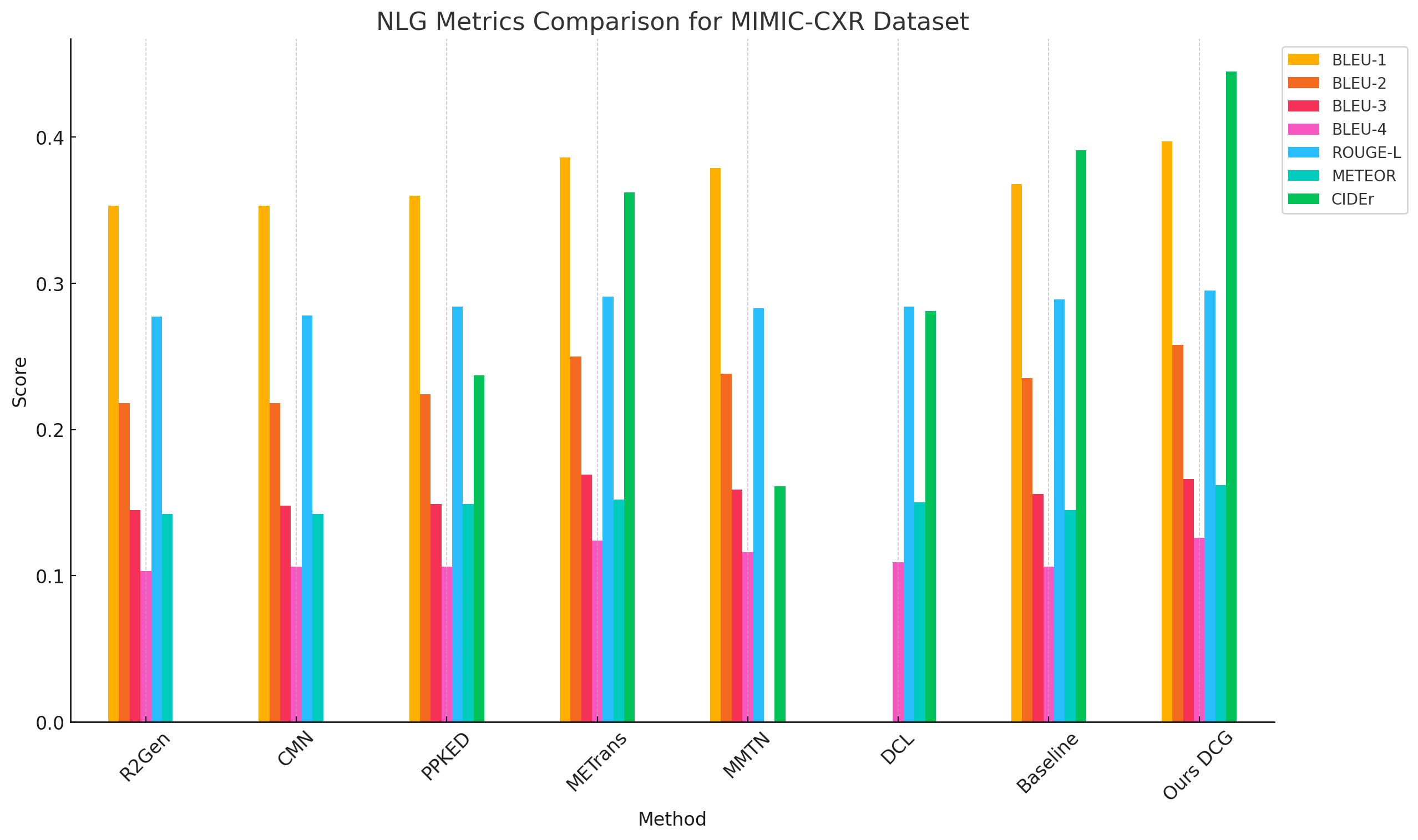}
        \caption{MIMIC-CXR NLG Metrics Comparison}
        \label{fig:mimic_cxr_nlg}
    \end{subfigure}
    \hfill
    \begin{subfigure}[b]{0.27\textwidth}
        \centering
        \includegraphics[width=\textwidth]{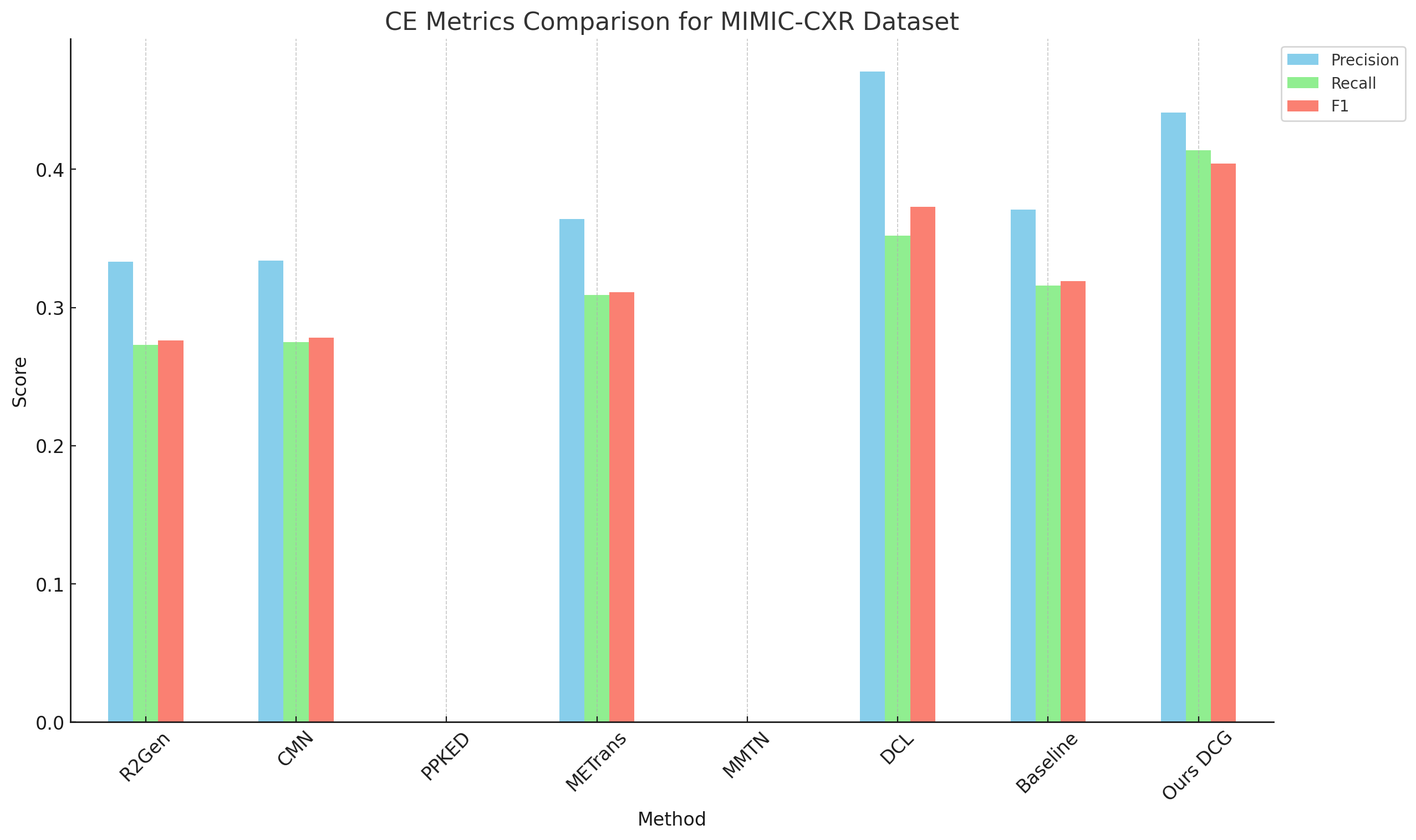}
        \caption{MIMIC-CXR CE Metrics Comparison}
        \label{fig:mimic_cxr_ce}
    \end{subfigure}
    
    % Row 2: IU-Xray radar chart and MIMIC-CXR radar chart
    \begin{subfigure}[b]{0.27\textwidth}
        \centering
        \includegraphics[width=\textwidth]{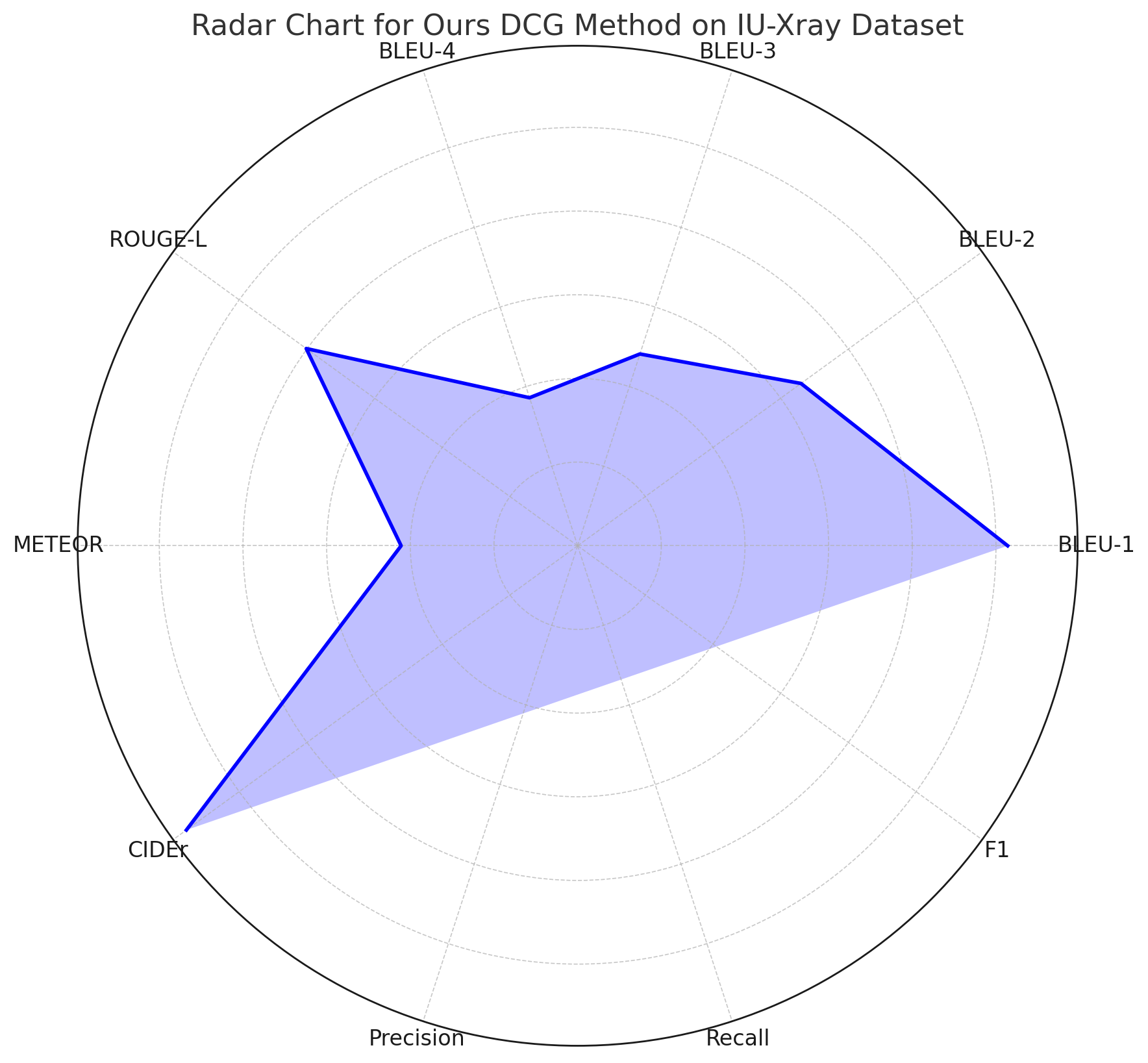}
        \caption{IU-Xray Radar Chart for Ours FODA-PG}
        \label{fig:iu_xray_radar}
    \end{subfigure}
    \hfill
    \begin{subfigure}[b]{0.27\textwidth}
        \centering
        \includegraphics[width=\textwidth]{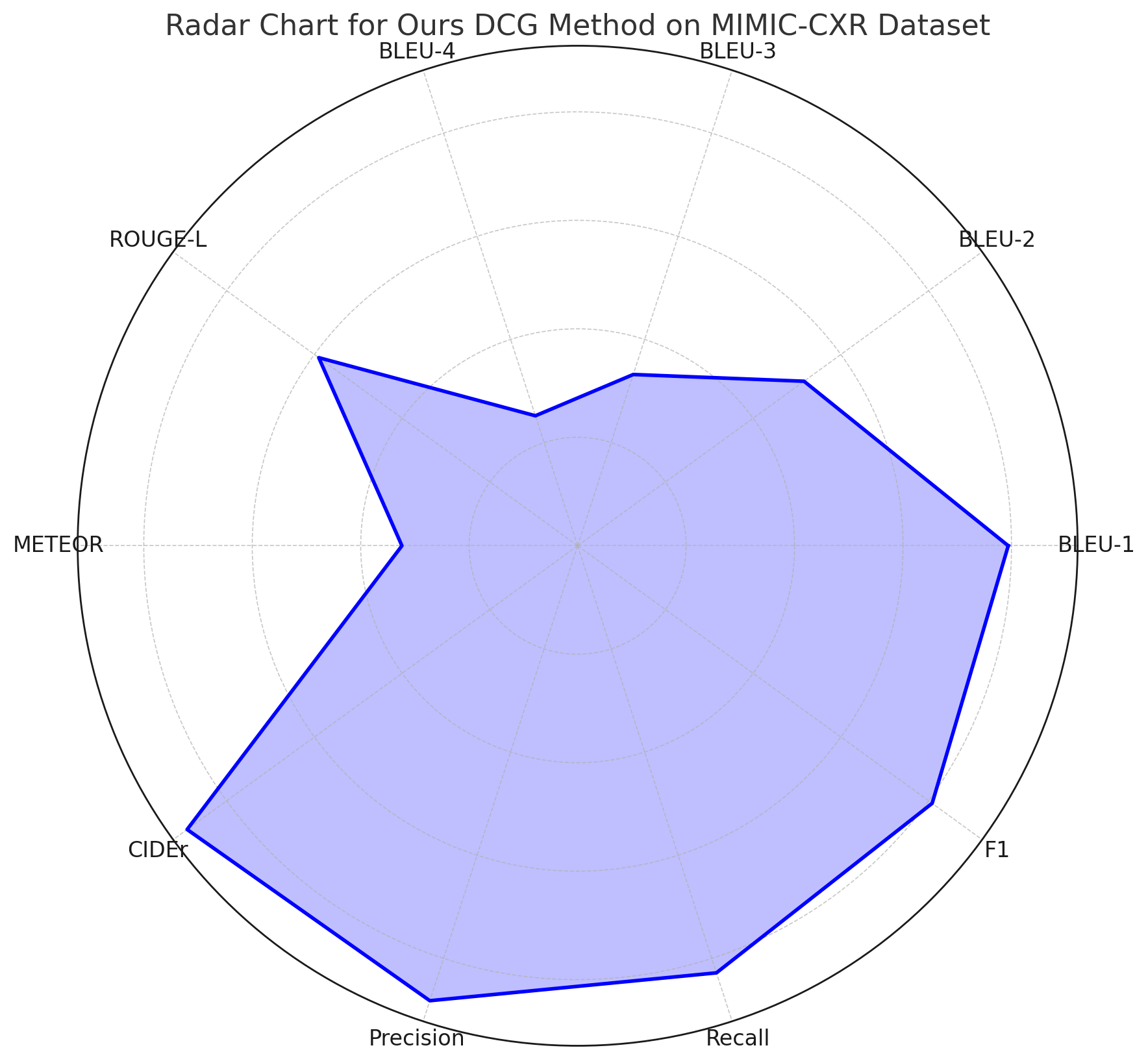}
        \caption{MIMIC-CXR Radar Chart for Ours FODA-PG}
        \label{fig:mimic_cxr_radar}
    \end{subfigure}

    \caption{Evaluating Natural Language Generation and Clinical Efficacy Metrics for Multiple Techniques across Radiography Datasets.}
    \label{tab:sota_comparison}
\end{figure}

\section{Experiment}
\subsection{Dataset}
We conducted an evaluation of our Fine-grained Organ-Disease Adaptive Partitioning Graph (FODA-PG) model using two established radiology reporting benchmarks: IU-Xray \cite{demner2015preparing} and MIMIC-CXR \cite{johnson2019mimic}, with preprocessing and dataset division protocols modeled after \cite{chen2020generating} to ensure a standardized comparison.

% In the development of our organ-disease graph, we adhered to the organ classifications as outlined in \cite{chen2020generating}. Our approach diverges in the granularity of disease identification, where instead of approximating about 20 disease findings related to specific organs or tissues, we distinguish meticulously between disease-specific and disease-free entities, resulting in a more complex and comprehensive graph. This detailed differentiation increases the total number of nodes in the graph, with IU-Xray featuring 191 nodes and MIMIC-CXR containing 276 nodes.

\subsection{Execution Configuration}
\subsubsection{Visual Feature Extractor}
In a departure from earlier methodologies that utilized ResNet-101 or DenseNet-121 trained on ImageNet for image encoding \cite{chen2022cross,chen2020generating,huang2023kiut}, we employ the Vision Transformer (ViT) from MedSAM \cite{ma2023segment} as our image encoder, specifically omitting the MLP neck to focus on extracting patch embeddings. With the ViT, an input image of 256*256 is transformed into a 16*16*768 feature map, which is subsequently reshaped into 256*768 patch embeddings. Consistent with established protocols \cite{li2023dynamic,chen2020generating}, our process involves handling paired images for the IU-Xray dataset and a single image for MIMIC-CXR. To standardize the output across different datasets, we reduce the number of patch embeddings from 1024 to 256.

\subsubsection{Graph Construction}
For the creation of our graph, we selectively use reports from the most closely matched images, identified via cosine similarity measures employed by BioMedCLIP \cite{zhang2023biomedclip}. As detailed in \cite{wang2023rethinking}, the reports are first segmented and preprocessed, and then a predefined list of organs and diseases are extracted through string matching using the Natural Language Toolkit (NLTK) \cite{bird2009natural}. Distinctions between disease-specific and disease-free cases are made by detecting terms like "no" and "normal" within the text. The DistilGPT2 model \cite{sanh2019distilbert}, with its Language Model (LM) Head removed, is utilized to derive node embeddings for all identified disease states, maintaining a dimensionality of 768.

\subsubsection{Text Decoder and Generation}
DistilGPT2 continues to serve as the text decoder within our framework. Our vocabulary is enriched with DistilGPT2’s tokens, along with additional [BOS] and [EOS] tokens to facilitate text generation. Following the standardization approach of previous CXR report generation models like that of Chen et al. \cite{chen2020generating}, we limit reports to 128 words, transform all text to lowercase, exclude special characters, and replace less common words with a placeholder token. 

% During the testing phase, the decoder can generate up to 256 subwords. A beam search algorithm, with a beam width of four, is implemented for generating reports, and a beam width of one is used during validation for a greedy search approach.

\subsubsection{Optimizing Parameters}
The training regimen involves the use of 8 NVIDIA A100 GPUs, supporting a batch size of 32 for a total of 30 epochs across both datasets. We select the training checkpoint that achieves the highest CIEDr score for final evaluations. Initial learning rates are set at 5e-6 for the encoder and 5e-5 for other parameters, with all other AdamW hyperparameters remaining at their default settings.

% \begin{table}[t]
% \centering
% \scalebox{0.9}{
% \begin{tabular}{ccccccc}
% \toprule
% \multirow{2}{*}{Settings} & 
% \multicolumn{2}{c}{Image Encoder} & 
% \multicolumn{1}{c}{Size} &
% \multicolumn{3}{c}{NLG Metric} \\ 
% \cmidrule{2-7}
% & Model & Pretrained & &
% BLEU-4 & ROUGE-L & CIDEr \\ 
% \midrule
% (a) & ViT-B/16 & BioMedCLIP \cite{zhang2023biomedclip} & 224 & 0.163 & 0.378 & 0.422 \\
% (b) & CvT & ImageNet21k \cite{dosovitskiy2020image} & 384 & 0.165 & 0.379 & 0.426 \\
% (c) & ViT-B/16 & MedSAM \cite{ma2023segment} & 512 & 0.170 & 0.392 & 0.537 \\
% \bottomrule
% \end{tabular}
% }
% \caption{Ablation study of the visual encoder. (a) is the Vision Transformer (ViT) pretrained on BioMedCLIP \cite{zhang2023biomedclip}; (b) is the Convolutional Vision Transformer (CvT) pretrained on ImageNet-21K; (c) is the ViT fine-tuned on medical image segmentation using MedSAM \cite{ma2023segment}.}
% \label{tab:ablation_visual_encoder}
% \end{table}
\begin{figure}
    \centering
    \includegraphics[width=0.35\textwidth]{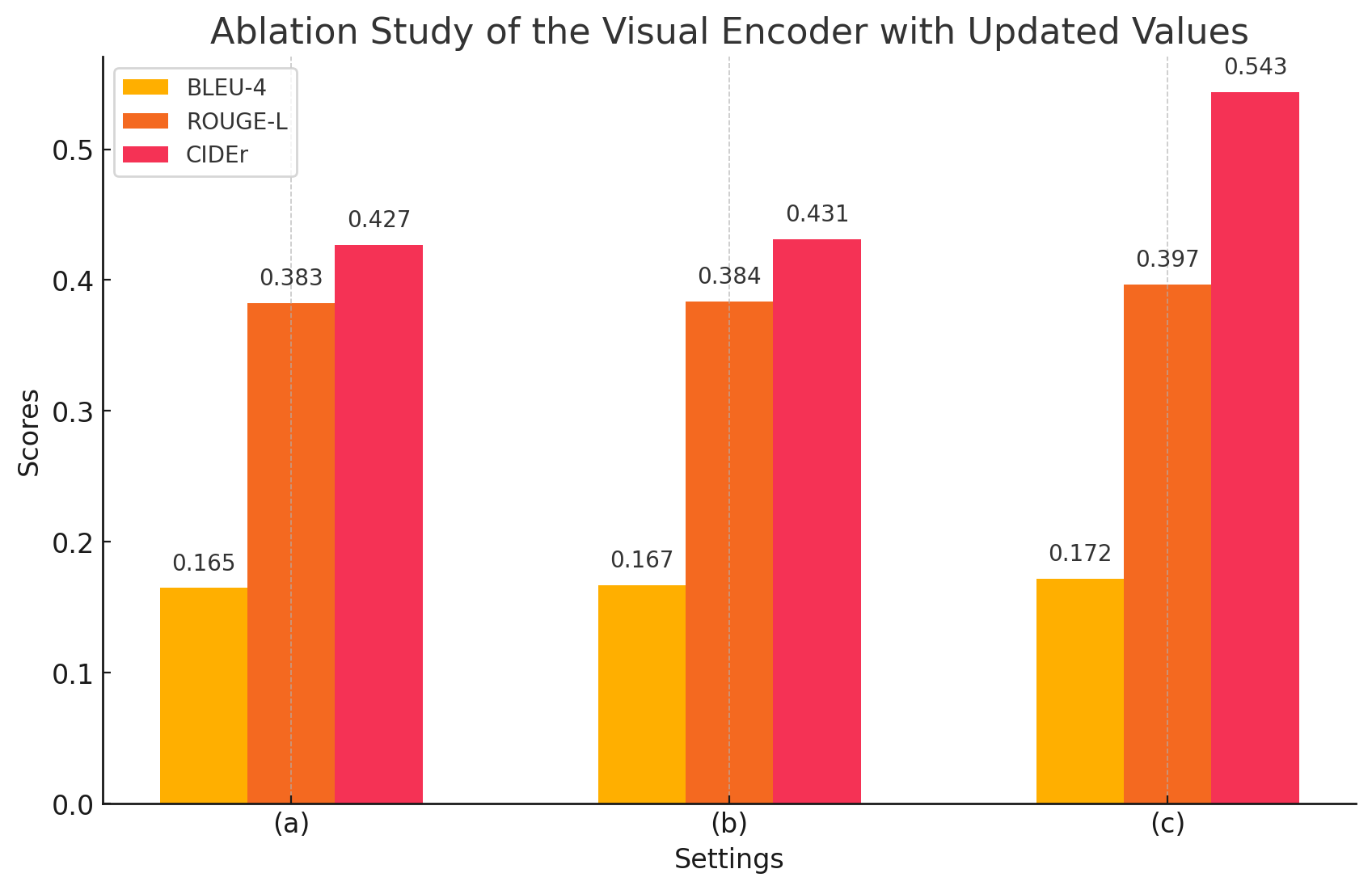}
    \caption{Assessing Updated Visual Encoder Setups: (a) BioMedCLIP-pretrained ViT \cite{zhang2023biomedclip}; (b) ImageNet-21K-pretrained CvT; (c) MedSAM-fine-tuned ViT for Medical Image Segmentation \cite{ma2023segment}.}
    \label{tab:ablation_visual_encoder}
\end{figure}

\subsection{Evaluation Metrics}

Our performance evaluation employs a comprehensive set of Natural Language Generation (NLG) metrics, including CIDEr \cite{vedantam2014cider}, BLEU \cite{papineni2002bleu}, ROUGE-L \cite{lin2004rouge}, and METEOR \cite{banerjee2005meteor}, complemented by Clinical Efficacy (CE) metrics. 

% BLEU is utilized to measure the n-gram overlap between the generated and reference texts, although it may be susceptible to textual biases. In contrast, CIDEr enhances the evaluation of medical report generation (MRG) systems by emphasizing the relevance of topic-specific terms. For clinical accuracy, we deploy the CheXPert tool \cite{smit2020chexbert} to annotate reports with more than 10 medical terminologies, evaluating the clinical pertinence through F1-Score, Precision, and Recall metrics. Notably, the application of CE metrics is confined to the MIMIC-CXR dataset \cite{johnson2019mimic}, given that IU-Xray lacks integration with the CheXPert labeling framework.

\section{Experiment Results}
\subsection{Comparison with Baselines}
To validate the superiority of our approach, we benchmarked our model, termed FODA-PG, against leading models in the domain of Medical Imaging Narrative Generation (ING) using the established IU-Xray and MIMIC-CXR datasets, as detailed in Figure \ref{tab:sota_comparison}. Among the models evaluated were R2Gen \cite{chen2020generating}, the foundational model for ING; CMN \cite{chen2022cross} and PPKED \cite{liu2021exploring}, which incorporate organ-disease knowledge graphs; and the more recent METrans \cite{wang2023metransformer}, MMTN \cite{cao2023mmtn}, and DCL \cite{li2023dynamic}. Our model demonstrated superior performance across both Natural Language Generation (NLG) and Clinical Effectiveness (CE) metrics. The BLEU \cite{papineni2002bleu} score quantifies the n-gram similarity between the generated and reference reports, while ROUGE-L \cite{lin2004rouge} assesses the longest contiguous matching sequence of words, and METEOR \cite{banerjee2005meteor} evaluates alignment at a more granular level, factoring in synonymy and paraphrasing. Importantly, an elevated CIDEr score reflects the semantic depth and clinical relevance of the reports crafted by our method.

% \begin{table}[t]
% \centering
% \scalebox{0.9}{
% \begin{tabular}{cccc|ccc}
% \toprule
% \multirow{2}{*}{Settings} &
% \multicolumn{2}{c}{Node} & 
% \multirow{2}{*}{Information Fusion} &
% \multicolumn{3}{c}{NLG Metric} \\ 
% \cmidrule{2-3} \cmidrule{5-7}
% & Encoder & GCN & & 
% BLEU-4 & ROUGE-L & CIDEr \\
% \midrule
% Baseline & - & - & - & 0.170 & 0.392 & 0.537 \\
% \midrule
% (a) & PubMedBERT \cite{gu2020domain} & \checkmark & \checkmark & 0.165 & 0.376 & 0.426 \\
% (b) & DistilGPT2 & - & \checkmark & 0.170 & 0.379 & 0.490 \\
% (c) & DistilGPT2 & \checkmark & - & 0.172 & 0.395 & 0.509 \\
% (d) & DistilGPT2 & \checkmark & \checkmark & \textbf{0.186} & \textbf{0.401} & \textbf{0.578} \\
% \bottomrule
% \end{tabular}
% }
% \caption{Ablation study of the node encoder and information fusion method.}
% \label{tab:ablation_node_encoder}
% \end{table}
\begin{figure}
    \centering
    \includegraphics[width=0.35\textwidth]{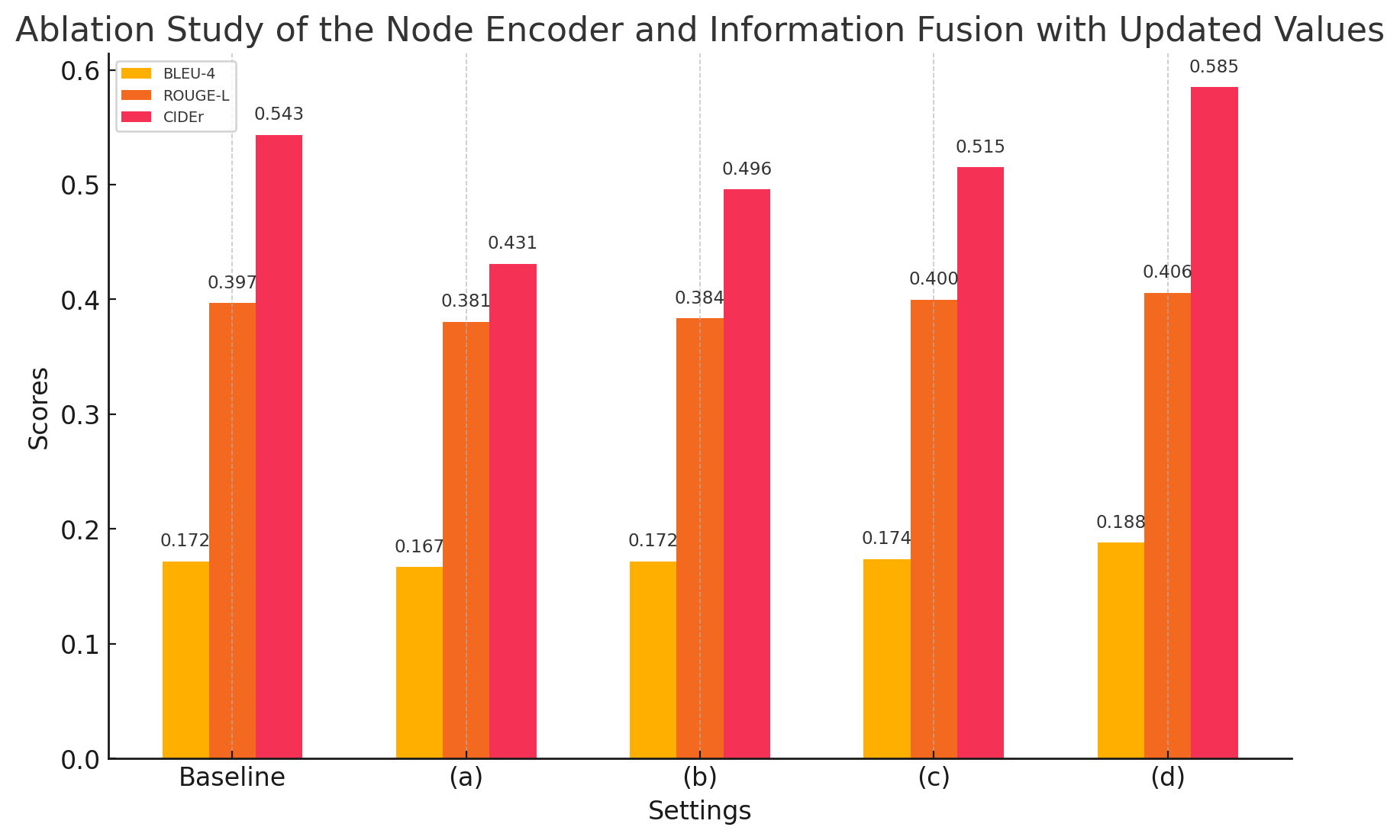}
    \caption{Node Representation and Multi-Source Integration Ablation Analysis with Revised Configurations.}
    \label{tab:ablation_node_encoder}
\end{figure}

\subsection{Ablation Study}
In this subsection, we delineate the frequency and types of normal and abnormal disease manifestations within the IU-Xray and MIMIC-CXR datasets, underscoring the importance of differentiating between disease-specific and disease-neutral categories. This is followed by an evaluation of offline image retrieval performance using BioMedCLIP \cite{zhang2023biomedclip}, and an exploration of the individual contributions of each component within our Fine-grained Organ-Disease Adaptive Partitioning Graph (FODA-PG) model. The effects of different visual encoders on model accuracy are presented in Table \ref{tab:ablation_visual_encoder}, and the impacts of various node encoders, node modeling techniques, and information fusion strategies are depicted in Figure \ref{tab:ablation_node_encoder}.

\subsubsection{Dataset Distributions} 
Analysis of the disease entities extracted from the reports shows a higher prevalence of disease-neutral entities in IU-Xray compared to disease-specific ones, with a more balanced distribution in MIMIC-CXR. This balance is attributed to our methodological refinement of sentence segmentation, such as the parsing of phrases like "No pneumothorax, pleural effusion, or focal air space consolidation". Disease-specific entities, including "pneumothorax" and "effusion", show a long-tailed frequency distribution, which is typical for clinical datasets.

\subsubsection{Retrieval Performance} 
Our validation of the FODA-PG-enhanced methodology involved assessing the alignment between retrieved and actual reports via BioMedCLIP \cite{zhang2023biomedclip}, utilizing predefined pairs of disease-specific and disease-neutral entities. Notably, increasing the number of retrieved images improved the recall of disease entities, albeit with a slight reduction in precision, reaching over 51\% entity recall when retrieving three images.

\subsubsection{Visual Encoder} 
The efficacy of Medical Imaging Narrative generation hinges significantly on the quality of visual representations. We evaluated several top-tier image encoders tailored to both medical and general imagery, as outlined in Table \ref{tab:ablation_visual_encoder}. The performance metrics were closely matched between ViT-B/16@224, initialized with BioMedCLIP \cite{zhang2023biomedclip}, and CvT@384 pretrained on ImageNet21k. Notably, MedSAM \cite{ma2023segment}, which focuses on medical imagery, demonstrated superior performance, underscoring the importance of fine-grained region-of-interest (ROI) features in medical diagnostics.

\subsubsection{Vertex Representation and Multi-Source Integration} 
The text-based construction of our disease graph prompted the use of text encoders for node embedding. In contrast to previous methods using SciBERT \cite{li2023dynamic}, our approach included trials with PubMedBERT aligned with BioMedCLIP \cite{zhang2023biomedclip} for enhanced multi-modal integration. Despite introducing graph priors, this adaptation did not improve report generation, potentially due to the limited size of the IU-Xray dataset, which may hinder effective learning of correlations between PubMedBERT's node embeddings and DistilGPT2's token embeddings. The configurations (b) and (c) explored the utility of graph convolutional networks and multi-head cross-attention mechanisms, respectively, in enhancing node and patch embedding interactions. Our final configuration (d) combined these elements to optimize the generation of clinically relevant reports.

\section{Conclusion and Discussion}
In this study, we introduce a pioneering method for constructing organ-disease graphs to enhance the generation of Medical Imaging Narratives. Traditional approaches often restrict their focus to a narrow spectrum of diseases and fail to capture the nuanced distinction between normal and pathological findings as comprehensively as actual clinical narratives do. Our proposed Fine-grained Organ-Disease Adaptive Partitioning Graph (FODA-PG) method leverages similarity-based retrieval to meticulously construct fine-grained organ-disease graphs. This approach meticulously categorizes nodes into disease-specific or disease-neutral categories, reflecting their pathological significance or absence thereof. Rigorous testing on established benchmarks like IU-Xray and MIMIC-CXR substantiates the robustness and accuracy of our approach.

% However, the current implementation of FODA-PG is not without limitations. Notably, the retrieval of reports is performed offline and is not optimized specifically for the intricacies of chest X-ray analysis, which could enhance specificity and accuracy. Future developments will focus on refining these aspects to improve the real-time applicability and precision of the model. Our ongoing efforts aim to refine these mechanisms, enhancing the utility and accuracy of Medical Imaging Narrative automation in clinical settings.

\vspace{12pt}

\end{document}